\documentclass[sigconf]{acmart}

\setcopyright{none}

\copyrightyear{2023}
\acmYear{2023}
\setcopyright{acmcopyright}

\copyrightyear{2024}
\acmYear{2024}
\setcopyright{acmlicensed}\acmConference[WSDM '24]{Proceedings of the 17th ACM International Conference on Web Search and Data Mining}{March 4--8, 2024}{Merida, Mexico}
\acmBooktitle{Proceedings of the 17th ACM International Conference on Web Search and Data Mining (WSDM '24), March 4--8, 2024, Merida, Mexico}
\acmPrice{15.00}
\acmDOI{10.1145/3616855.3635786}
\acmISBN{979-8-4007-0371-3/24/03}
\usepackage[utf8]{inputenc}
\usepackage{longtable}
\usepackage{threeparttable}
\usepackage{xspace}
\usepackage{appendix}
\usepackage{booktabs}
\usepackage{subcaption}
\usepackage{textcomp}
\usepackage{amsmath,amsthm}
\usepackage{enumitem}
\usepackage{algorithm}
\usepackage[noend]{algpseudocode}
\algtext*{EndProcedure}%
\usepackage{xcolor,colortbl}
\usepackage{footnote}
\usepackage{url}
\usepackage{tabularx}
\usepackage{multicol}
\usepackage{multirow}
\usepackage{pifont}
\usepackage{bm}
\usepackage{listings}
\usepackage{wrapfig}
\usepackage{relsize}
\usepackage{mathtools}
\usepackage{bbm}
\usepackage{blindtext}
\usepackage{balance}
\usepackage{mdwlist}
\usepackage{derivative}
\usepackage{bbold}
\usepackage{array}
\usepackage{makecell}
\usepackage{verbatim}
\setlist[enumerate]{leftmargin=*}
\setlist[itemize]{leftmargin=*}

\usepackage{wasysym}
\usepackage{pifont}%

\definecolor{green}{rgb}{0,0.7,0.3}

\newcommand{\matX}{\mathbf{X}}

\newcommand{\cmark}{\ding{51}}%
\newcommand{\xmark}{\ding{55}}%

\newcommand{\method}{\text{\textsc{SpotTarget}}\xspace}

\theoremstyle{plain}
\newtheorem*{example}{Illustrative Example}

\definecolor{msred}{rgb}{0.753,0.314,0.302}

\definecolor{lblue}{rgb}{0.9411, 0.9725, 1.0}
\definecolor{llblue}{rgb}{0.3, 0.6, 1.0}
\definecolor{lgreen}{rgb}{0.908, 0.961, 0.908}
\definecolor{lblue}{rgb}{0.94, 0.96, 1.0}
\definecolor{lred}{rgb}{1.0, 0.885, 0.885}
\definecolor{lorange}{rgb}{1.0, 0.88, 0.69}

\newcolumntype{L}[1]{>{\raggedright\arraybackslash}p{#1}}
\newcolumntype{C}[1]{>{\centering\arraybackslash}p{#1}}

\newcolumntype{S}[1]{>{\columncolor{lblue}\centering\arraybackslash}p{#1}}
\newcolumntype{D}[1]{>{\columncolor{lgreen}\centering\arraybackslash}p{#1}}
\newcolumntype{R}[1]{>{\columncolor{lred}\centering\arraybackslash}p{#1}}

\newcolumntype{O}[1]{>{\columncolor{lorange}\centering\arraybackslash}p{#1}}

\theoremstyle{plain}
\newtheorem{theorem}{Theorem}
\definecolor{Gray}{gray}{0.85}
\newtheorem{observation}{Observation}

\usepackage{listings}
\usepackage{color}

\definecolor{dkgreen}{rgb}{0,0.6,0}
\definecolor{ggray}{rgb}{0.5,0.5,0.5}
\definecolor{mauve}{rgb}{0.58,0,0.82}

\author{Jing Zhu$^*$}
\affiliation{\institution{University of Michigan, Ann Arbor}}
\email{jingzhuu@umich.edu}

\author{Yuhang Zhou$^*$}
\affiliation{\institution{University of Maryland, College Park}}
\email{tonyzhou@umd.edu}

\author{Vassilis N. Ioannidis}
\affiliation{\institution{AWS AI Research and Education}}
\email{ivasilei@amazon.com}

\author{Shengyi Qian}
\affiliation{\institution{University of Michigan, Ann Arbor}}
\email{syqian@umich.edu}

\author{Wei Ai}
\affiliation{\institution{University of Maryland, College Park}}
\email{aiwei@umd.edu}

\author{Xiang Song}
\affiliation{\institution{AWS AI Research and Education}}
\email{xiangsx@amazon.com}

\author{Danai Koutra}
\affiliation{\institution{University of Michigan, Ann Arbor}}
\email{dkoutra@umich.edu}
\renewcommand{\shortauthors}{Jing Zhu et al.}

\begin{document}

\begin{abstract}

While Graph Neural Networks (GNNs) are remarkably successful in a variety of high-impact applications, 
we demonstrate that, in link prediction, the common practices of including the edges being predicted in the graph at training and/or test have outsized impact on the performance of low-degree nodes.
We theoretically and empirically investigate how these practices impact node-level performance across different degrees.  
Specifically, we explore three issues that arise: 
(I1) \textit{overfitting}; 
(I2) \textit{distribution shift}; and 
(I3) implicit \textit{test leakage}. 
The former two issues lead to poor generalizability to the test data, while the latter leads to overestimation of the model's performance and directly impacts the deployment of GNNs.  
To address these issues in a systematic way, we introduce an effective and efficient GNN training framework, \method,  
which leverages our insight on low-degree nodes: (1) at training time, it excludes a (training) edge to be predicted if it is incident to at least one low-degree node; and (2) at test time, it excludes \textit{all} test edges to be predicted (thus, mimicking real scenarios of using GNNs, where the test data is not included in the graph). 
\method helps researchers and practitioners adhere to best practices for learning from graph data, which are frequently overlooked even by the most widely-used frameworks. 
Our experiments on various real-world datasets show that \method makes GNNs {up to} {15$\times$} more accurate in sparse graphs, and significantly improves their performance for low-degree nodes in dense graphs.

 \end{abstract}



\title{Pitfalls in Link Prediction with Graph Neural Networks: Understanding the Impact of Target-link Inclusion \\
\& Better Practices} 
\renewcommand{\shorttitle}{Pitfalls in Link Prediction with Graph Neural Networks} 
\maketitle

{\let\thefootnote\relax\footnotetext{{$^*$ equal contribution}}}
\section{Introduction}
\label{sec:intro}
\begin{figure*}[t!]
    \centering
    \includegraphics[width=.65\linewidth]{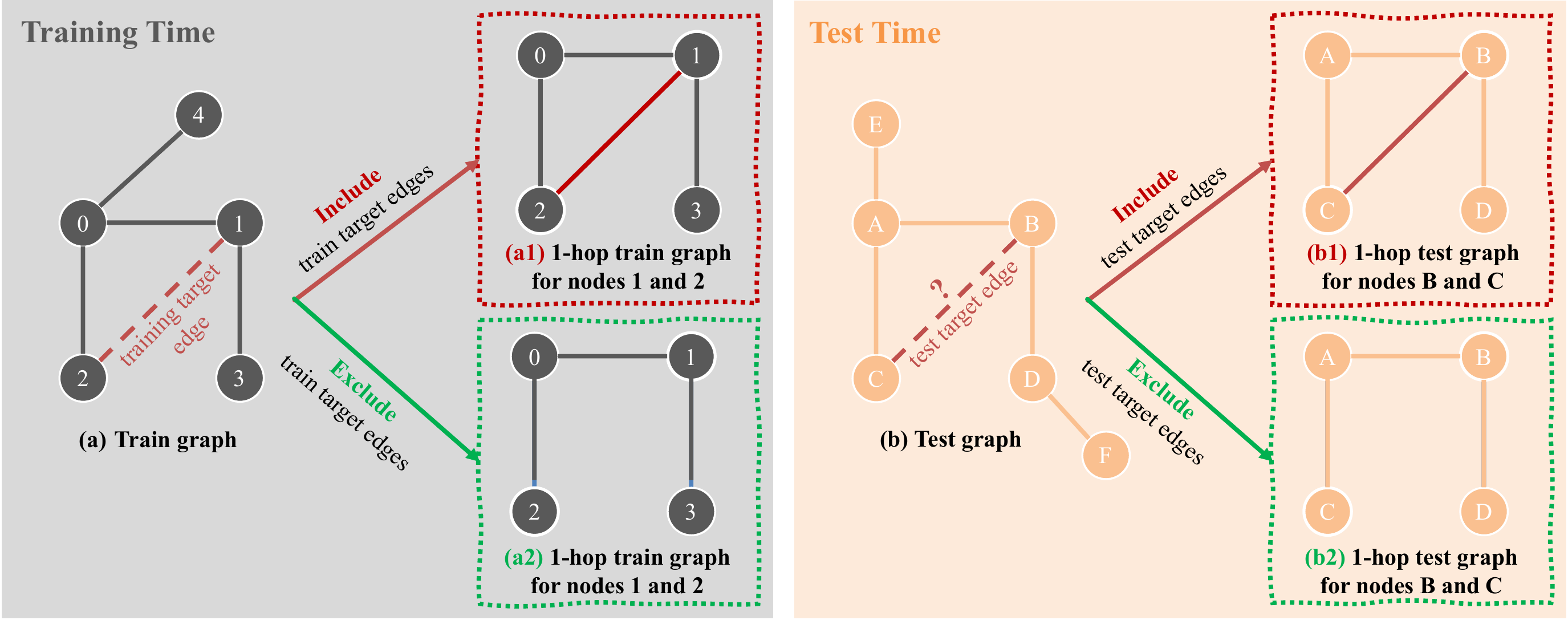}
    \vspace{-0.35cm}
    \small
    \begin{tabular}{cccc}
\toprule
 \rotatebox[origin=c]{0}{\textbf{\bf Training Target Edges }}
  & \rotatebox[origin=c]{0}{\textbf{\bf Test Target Edges }}
  &  \rotatebox[origin=c]{0}{\textbf{\bf Results }} 
  &  \rotatebox[origin=c]{0}{\textbf{\bf Issues }} \\  
  \hline 
  Include \textcolor{red}{(a1)} & - & \xmark & (I1) Overfitting\\
  Include \textcolor{red}{(a1)} & Exclude \textcolor{green}{(b2)} & \xmark & (I2) Distribution shift  \\ 
  - & Include \textcolor{red}{(b1)} & \xmark & (I3) Leakage \\
  Exclude \textcolor{green}{(a2)} & Exclude \textcolor{green}{(b2)} & \color{green}{\cmark} & -- \\ 
  
\bottomrule
\end{tabular}
    \caption{
    The pitfalls of including  target links  as message-passing edges during training or test time, and the issues that arise from these practices. 
    [Left] Training time: Given a toy train graph and training target edge $e_{12}$ in (a), we illustrate the impact of the inclusion (a1) and exclusion (a2) of $e_{12}$ on the 1-hop induced train graph for nodes 1 and 2, which is used for message passing. 
    [Right] Test time: We give the same illustration for a test graph and test target edge $e_{BC}$ in (b).  
    [Table] Overview of the three main issues and when they arise: (I1) When including train targets, GNNs \textit{overfit} them instead of making predictions based on the graph structure and node features. 
     (I2) When train target links are present but test target edges are absent, there is a \textit{distribution shift} between training and testing. 
     (I3) The presence of test target links causes \textit{implicit test leakage}. 
    }
    \label{fig:teaser}
    \vspace{-0.3cm}
\end{figure*}

Graphs or networks are key representations for relational data that occur in many %
scientific and industrial applications. 
Link prediction, the task of predicting whether a link is likely to form between two nodes or entities in a graph,  
has many downstream applications such as drug repurposing, recommendation systems, and knowledge graph completion~\cite{liben2003link, adamic2003friends, koren2009matrix, bordes2013translating, zeng2020repurpose, martinez2016survey}. 
It is also widely used as a pre-training method to produce high-quality entity representations that can be used in various business applications ~\cite{hu2020gpt, hu2019strategies}. 
Techniques to solve this task range from heuristics--e.g., predicting links based on the number of common neighbors between a pair of nodes--to graph neural network (GNN) models , which rely on message passing and leverage both the graph structure and node features. In recent years, GNN-based methods, which formulate the link prediction problem as a binary classification problem over node pairs, have led to state-of-the-art performance in many high-impact applications and have become the go-to approach both in industrial settings and academia~\cite{zhang2018link, kipf2016variational, zhang2021labeling, ioannidis2022efficient}.

In this work, we focus on key pitfalls when training GNN models for the link prediction task, which we have found to cause significant disparities in node-level performance.  
Specifically, we investigate the common practices of including in the graph the \textit{target links} (i.e., the edges for which the existence or absence is being predicted) 
at training and/or test time, and considering them during message passing~\cite{dong2022fakeedge, zhang2018link}.  
The inclusion of (training) target links at training time leads to two major issues, \textit{overfitting} (I1) and \textit{distribution shift} (I2), while the inclusion of target edges in the test graph data causes \textit{implicit data leakage} (I3) through neighborhood aggregation. 
In turn, these issues lead to poor performance for GNN models and inability to effectively generalize to (truly) unobserved links at test time. 
We give an illustrative example of the issues in message passing in Fig.~\ref{fig:teaser}.

\begin{example} 
In Fig.~\ref{fig:teaser}(a), $e_{12}$ is a training target link for which we want to predict the existence. 
When this edge is not excluded from the training graph, GNNs would use the message-passing graph shown in Fig.~\ref{fig:teaser}(a1) for nodes $1$ and $2$, which leads to overfitting on $e_{12}$ and memorizing its existence instead of learning to predict it based on the graph structure and node features.
Moreover, in a realistic testing scenario as in Fig.~\ref{fig:teaser}(b) where the goal is to predict whether the edge $e_{BC}$ exists or not, GNNs would use the message-passing graph shown in Fig.~\ref{fig:teaser}(b2) for nodes $B$ and $C$, where the two nodes are disconnected. 
This leads to distribution shift: there is discrepancy between the message-passing graphs used during training and testing despite the similarity between %
the target links.

On the other hand, at test time, including the test target links in the test graph (e.g., edge $e_{BC}$ in Fig.~\ref{fig:teaser}(b1)) causes data leakage. In our example, during neighborhood aggregation,  the target node $B$ would aggregate the messages from $C$ and vice versa, resulting in a higher likelihood of predicting the existence of edge $e_{BC}$ compared to the case where the link does not exist in the message-passing graph. 
However, in real-world applications, the goal is to predict \emph{future} links that are \emph{not observed} in the data, so the inclusion of test target links corresponds to implicit data leakage. 
\end{example}

The pitfalls of including the target links in the graph at train and/or test time are commonplace in many GNN-based frameworks. 
For example, PyTorch Geometric (PyG)~\cite{pytorch-geometric}, a commonly-used repository, 
does not support excluding target edges when constructing the mini-batch graphs for training. 
Another popular library, DGL~\cite{dgl}, for the first four years of its existence, did not include the function of excluding training target links in the official code examples that have been used by numerous researchers and practitioners. 
The majority of papers fail to reference the exclusion of target links as a consideration in their empirical analyses, and, anecdotally, multiple authors with both industry and academic experience have observed that these pitfalls often occur in practice.  
Although there have been efforts to deliberately eliminate the test-time pitfall in some popular benchmarks~\cite{hu2020open}, 
it is still a commonly overlooked problem in applications that rely on proprietary data. 
Data contamination has also been a major issue in model evaluation, especially in the era of large language models\cite{zhou2023don, sainz2023nlp, golchin2023time, zhou2023explore}.

We demonstrate theoretically and empirically that low-degree nodes suffer more from the  
inclusion of target edges as it causes more significant relative degree changes for them compared to other nodes. 
Intuitively, for high-degree nodes, the target links that are erroneously considered have limited impact on the performance since they are only a small fraction of the edges considered during message passing.  
Thus, these practices significantly impact real-world applications, %
where the observed data are often incomplete and very sparse, with many low-degree nodes~\cite{reddy2022shopping, faloutsos1999power, leskovec2020mining,leskovec2005graphs}. 
To address the three issues (I1-I3), we introduce a GNN training framework, \method, to systematically and efficiently exclude the target links at training and test time, as well as check if target test edges are excluded for any user-defined dataset. 
Although excluding all training target links is an ideal solution, our analysis indicated that it significantly corrupts the mini-batch graph and renders learning with GNNs challenging. 
Our theoretical and empirical analysis shows that excluding the target links that are incident to at least one low-degree node achieves the best trade-off between avoiding the issues (I1, I2) and learning powerful node representations at training time. 
At test time, we argue that it is important to mimic real scenarios and avoid leakage for \textit{all} target edges by excluding them from the test graph. 
Our key contributions are:

\begin{itemize}
    \item \textbf{Systematic Analysis of the Target Link Inclusion Practices}: Focusing on link prediction, we perform the first thorough theoretical and empirical analysis on the effect of including target edges as message-passing edges at training and test time. Our key insight is that low-degree nodes tend to suffer more from the issues that arise from these pitfalls.   
    \item \textbf{Efficient Unified Framework}: We introduce the first unified GNN training framework, \method, which automatically tackles these issues at training and test time. %
    During training, for efficiency, \method leverages our theoretical insight and excludes target links incident to at least one low-degree node. At test time, it excludes all target edges. These strategies ensure 
    generalizable and robust model training without any data leakage issues. 
    \method is also easy-to-use and scalable, and 
     helps researchers and practitioners adhere to best practices, which are frequently overlooked even by the most widely-used GNN frameworks. We integrated it as a plug-and-play module in DGL. %
    \item \textbf{Extensive Experiments}: To quantify the effect of including the target links as message-passing edges during training and test time, we conduct extensive experiments on various datasets, spanning from commonly-used link prediction benchmarks to real-world datasets. 
    We show that \method makes GNN models up to {15$\times$} more accurate on sparse graphs, and significantly improves their performance for low-degree nodes on dense graphs. 
     
\end{itemize}

\section{Related Work} 
\label{sec:related}

\noindent \textbf{Link Prediction using GNNs.} Graph neural networks (GNNs) are popular neural network architectures that learn representations by capturing the interactions between objects. While perhaps most often used for node- or graph-level classication, the applications of GNNs have expanded to include edge-level inference tasks like link prediction. Methods that use GNNs for link prediction mainly fall into two categories: Graph Autoencoder (GAE)-based methods and enclosing subgraph-based methods. GAE-based methods use GNNs as the encoder of nodes, and edges are decoded by their nodes’ encoding vectors using score functions ~\cite{kipf2016variational, davidson2018hyperspherical, vashishth2019composition, zhu2021neural, you2019position, zhu2023touchup}. 
Enclosing subgraph-based methods, including SEAL ~\cite{zhang2018link, zhang2021labeling}, IGMC ~\cite{zhang2019inductive}, GraIL ~\cite{teru2020inductive}, TCL-GNN ~\cite{yan2021link}, first extract an enclosing subgraph for the target edge, then apply GNNs to encode the representations of the nodes in enclosing subgraph, and finally aggregate the node representations by pooling methods. The learned subgraph features are fed into a classifier to predict the existence or absence of the target edge. 
Even though enclosing subgraph-based methods such as SEAL give more accurate predictions, GAE-based methods are typically orders of magnitude faster to compute and require fewer computation resources. In real-world applications, graphs are often massive with many millions of nodes or even billions of nodes, so typically  GAE-based methods are employed~\cite{zheng2020distdgl}.

\vspace{0.05cm}
\noindent \textbf{Issues in Link Prediction using GNNs.} Unlike node classification where edges are solely used as message-passing edges, edges in link prediction have two separate roles: (1) message passing and (2) prediction objectives. 
This distinction is often overlooked; GNNs designed for node classification tasks are often adapted for link prediction by simply stacking a decoder function, without explicitly handling the message passing and target links separately. 
The training pitfalls caused by the existence of target edges were initially identified by SEAL~\cite{zhang2018link, wang2023neural}, which made efforts to mitigate them through negative injection. Building upon it, FakeEdge~\cite{dong2022fakeedge} discussed the distribution shift issue that occurs due to the presence of target links during training and the absence of target links at test time. 
They further proposed to always add or remove the target links, or combine the strategies for subgraph-based methods like SEAL. 
Unlike these works, we focus on performing a thorough and systematic analysis of all the issues caused by including target links at training and/or test time, and characterizing the disparate impact of these practices on the performance of nodes of varying degrees. 
Moreover, unlike SEAL and FakeEdge that only apply to subgraph-based models, our \method aims to %
systematically and efficiently address the issues for more scalable GAE-based models, which are commonly-used in real-world applications (e.g., web-scale recommender systems).  
For example, our training framework is 10x faster than FakeEdge (2 hours for one epoch vs.\ 20 hours on Ogbl-Citation2). %

\section{Preliminaries}
\label{sec:preliminaries}

In this section, we formally define key notions and the problem that we seek to solve. The major symbols we use are defined in Tab. ~\ref{tab:dfn}.

\begin{table}[]
\caption{Major symbols and their definitions.}
\vspace{-1.2em}
\centering
\resizebox{\columnwidth}{!}{
    \begin{tabular}{@{}l@{\hskip10pt}l}
    \toprule
    \textbf{Symbols} & \textbf{Definitions} \\
    \midrule
     $G$ & Graph \\
     $d_i$ & Degree of node $i$ \\
     $e_{ij}$ & The target edge between nodes $i, j$ to be predicted \\ %
      $T_{\text{tr}}$ & The set of train target edges \\
     $T_{\text{tst}}$ & The set of test target edges \\
     $T_{\text{low}}$ & Set of target edges incident to at least one low-degree node\\
     $\delta$ & Degree threshold to filter edges in $T_{\text{low}}$ \\
     \bottomrule
    \end{tabular}}
    \vspace{-1.7em}
    \label{tab:dfn}
\end{table}

\vspace{-0.2em}
\subsection{Definitions}

\noindent \textbf{Graphs.}
We consider a \textbf{graph} $G = (V,E,\matX)$, where $V$ is the set of vertices,
$E$ is the set of edges, and $\matX \in \mathbb{R}^{|V| \times d}$ represents the $d$-dimensional input node features. 
We denote as $N_{k}(u)$ the \textbf{$k$-hop neighbors} of node $u$, i.e., the set of nodes at a distance less than or equal to $k$ from $u$. 
The \textbf{degree} $d_u$ of node $u$ is defined as the number of its 1-hop neighbors or adjacent nodes, i.e., $d_u=|N_{1}(u)|$. 

\vspace{0.05cm}
\noindent \textbf{Link Prediction.} Given a graph $G = (V,E,\matX)$, the link prediction task aims to determine whether
there is or will be a link $e_{ij}$ between nodes $i$ and $j$, where $i, j \in V$ and $e_{ij} \notin E$. 
We refer to $e_{ij}$, the edge for which we want to predict the existence or absence, as \textbf{target edge or link}. 
We adopt the widely-used train-validate-test setting, where only the epoch that achieves best performance on validation links is evaluated on test edges.

In this paper, we distinguish different types of target links: 

\noindent (1)~\textbf{Training vs.\ test target links}: The training target edges, $T_{\text{tr}}$,  
are used to train a supervised link prediction model, while the test target links, $T_{\text{tst}}$, are the links for which we want to predict the existence or absence at test time (e.g., when evaluating the test performance or making predictions in real-world applications).  

\noindent (2)~\textbf{Target links that are incident to at least one low-degree node}: Based on our theoretical insights in Sec. \ref{sec:low-degree}, our framework leverages target links incident to at least one low-degree node (i.e., edges $e_{uv}$ for which $\min(d_u,d_v)$ is small), denoted as $T_{\text{low}}$.

\vspace{0.05cm}
\noindent \textbf{Graph Neural Networks.}
GNNs utilize a neighborhood aggregation scheme to learn a representation $h_v$ for each node $v$. Node representation is formulated as a $k$-round neighborhood aggregation schema:
{
$h_v^{(k)} = \text{COMBINE}^{(k)}(\{h_v^{(k-1)},  \text{AGGREGATE}^{(k)}(\{ h_u^{(k-1)}: u \in N_k(v) \}) \})$, 
\label{eq:gnn_formula}
}
{where AGGREGATE(.) is typically mean or max pooling, and COMBINE(.) can be a sum/concatenation/attention on nodes' ego- and neighbor-embeddings. 
}.
Given a set of target links, we define the \textbf{$k$-hop message-passing graph} of a GNN model as the induced subgraph that contains all the endpoint nodes of the target links, their k-hop neighbors, and the edges of the original graph that connect these nodes. Examples of (train and test) 1-hop message-passing graphs are given in Fig.~\ref{fig:teaser}.

\vspace{-0.2em}
\subsection{Problem Statement}
More formally, we tackle the following problem: 
Given a graph $G$, a link prediction task, and a base GNN model in a mini-batch training setting, 
we seek to: (1) investigate the issues that arise from the common practices of including the target links $T_{\text{tr}}$ and $T_{\text{tst}}$ as message-passing edges at training and test time, respectively, and (2) propose an efficient, unified and easy-to-use solution that automatically addresses these issues.

\section{Issues of Target-link Inclusion} %
\label{sec:issues}
In this section, we aim to explore the three issues that occur in link prediction with GNNs due to the practices of including the target links as message-passing edges at training and/or test time.

\subsection{Issues during Training Time}
\label{sec:train-pitfalls}
The presence of the training target links in the train graph data and their use as message-passing edges cause overfitting as well as  distribution shift.

\vspace{0.15cm}
\noindent \textbf{(I1) Overfitting}.
Suppose that we have an original train graph $G$, as shown in Fig.~\ref{fig:teaser}(a). GAE-based methods first generate node 1 and node 2's embeddings by aggregating their 1-hop neighbors' information and decode the likelihood of node 1 and node 2 forming an edge using a dot product decoder. When the target edge $e_{12}$ is present, node 1's embedding aggregates node 2's features, and vice versa. Since the training objective is to learn as high probability as possible for a link existing between node 1 and 2,  
GNNs would learn to overfit the training objective in order to predict the existence of the training target link. 
Similarly, subgraph-based models first find an enclosing subgraph for target edges $T_{\text{tr}}$ and then apply GNNs upon the enclosing subgraph to predict the link existence. When a target link is present in the enclosing subgraph as a message-passing edge, these models also suffer from overfitting issues. %
The overfitting issue leads to poor model generalizability to test data. 

\vspace{0.15cm}
\noindent \textbf{(I2) Distribution Shift}.
In typical GNN training processes for link prediction, the train target links $T_\text{tr}$ are present and used during message passing, while the test target links $T_\text{tst}$ are absent and never used during test. This practice poses a distribution shift problem. %
As an example, we consider the train graph in Fig.~\ref{fig:teaser}(a) along with $e_{12}$ as the train target link, and the test graph with $e_{BC}$ as the test target link in Fig.~\ref{fig:teaser}(b). As shown in Fig.~\ref{fig:teaser}(a1), at training time, when node 1 aggregates the messages from its neighbors,  node 2 is among its direct neighbors, and the message from node 2 contributes to the computation of node 1's embeddings. In a realistic test scenario (Fig.~\ref{fig:teaser}(b2)), future links are not observed in the test data; so, when node B aggregates the messages from its neighbors, it does not include any message from node C as the latter is not a direct neighbor. %
This poses a distribution shift between training and testing,  and also results in poor GNN model generalizability.

\subsection{Issues during Test Time}
\label{sec:leakage}

At test time, %
including the test targets links, $T_{\text{tst}}$, in message passing results in implicit data leakage.

\vspace{0.15cm}
\noindent \textbf{(I3) Data Leakage}.
As shown in Fig.~\ref{fig:teaser}(b1), when test target $e_{BC}$ exists in the test message-passing graph, the target node $B$ would aggregate messages from $C$ and vice versa, which results in a higher likelihood of predicting a link between nodes $B$ and $C$ during inference, compared to the case when $e_{BC}$ does not exist in the test graph in Fig.~\ref{fig:teaser}(b2). This leads to overestimation of the model's predictive performance and directly impacts the deployment of GNN models  since 
\emph{future} links that need to be predicted %
are never observed in real-world applications. 
 
\section{Proposed Framework: $\method$}
\label{sec:method}
\begin{figure}[t!]
    \centering
    \includegraphics[width=0.85\linewidth]{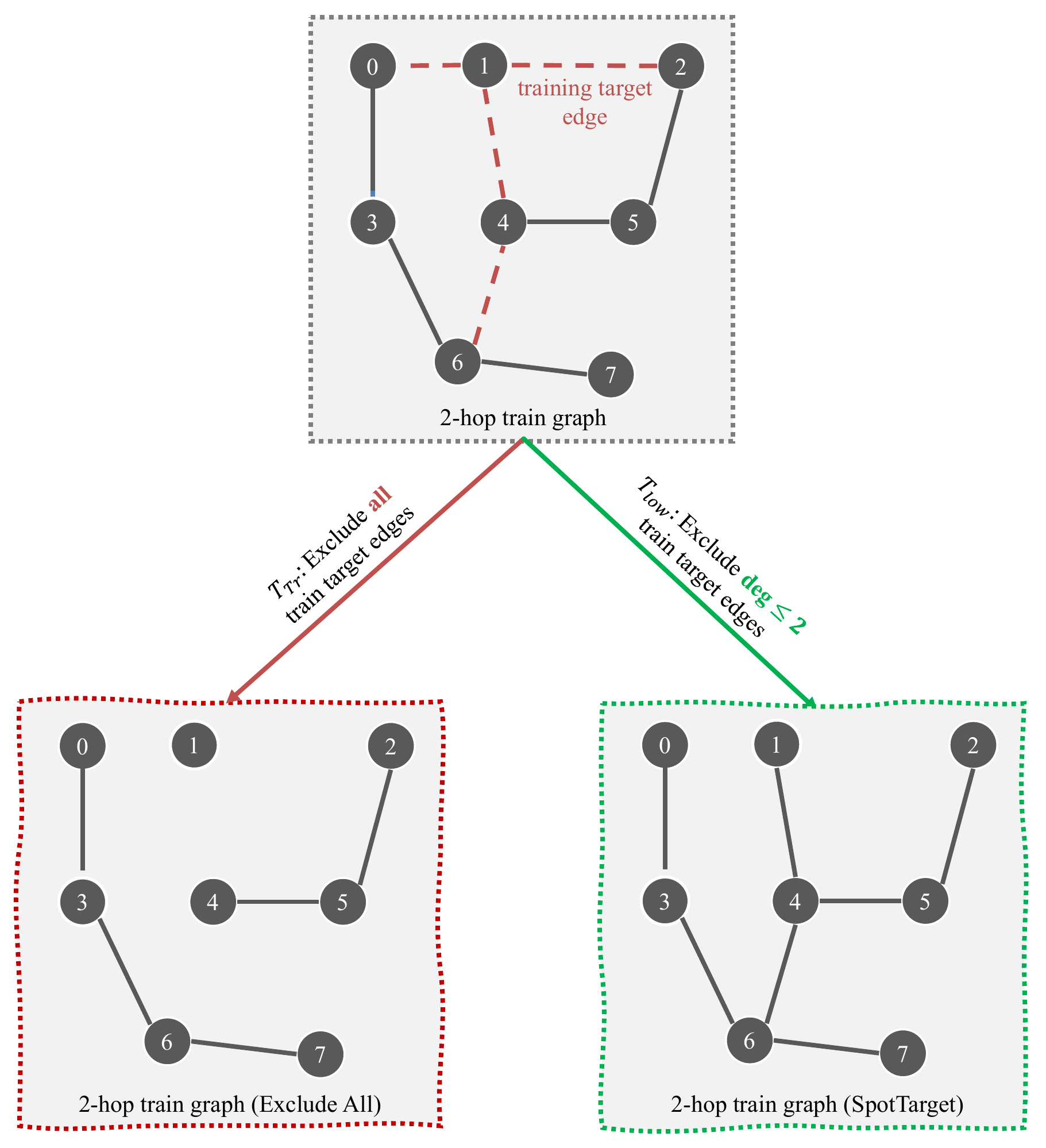}
    \vspace{-1.6em}
    \caption{Example 2-hop message-passing graph for a mini-batch of size 4. \textcolor{msred}{\bf Red lines} are train target links and black lines correspond to other message-passing edges induced by the target edges. As shown on the left, excluding all target (\textcolor{msred}{\bf red}) links $T_{\text{Tr}}$ during training results in three disconnected components. As shown on the right, if only edges incident to low-degree nodes are excluded $T_{\text{low}}$ (e.g., deg $\leq2$), the graph connectivity is preserved. Our proposed solution avoids significant corruption of the graph structure while simultaneously avoiding issues (I1) and (I2). }
    \vspace{-1.5em}
    \label{fig:2-hop graph}
\end{figure}

\begin{figure*}[t!]
\centering
    \begin{subfigure}{0.25\linewidth}
    \centering
    \includegraphics[width=0.85\columnwidth]{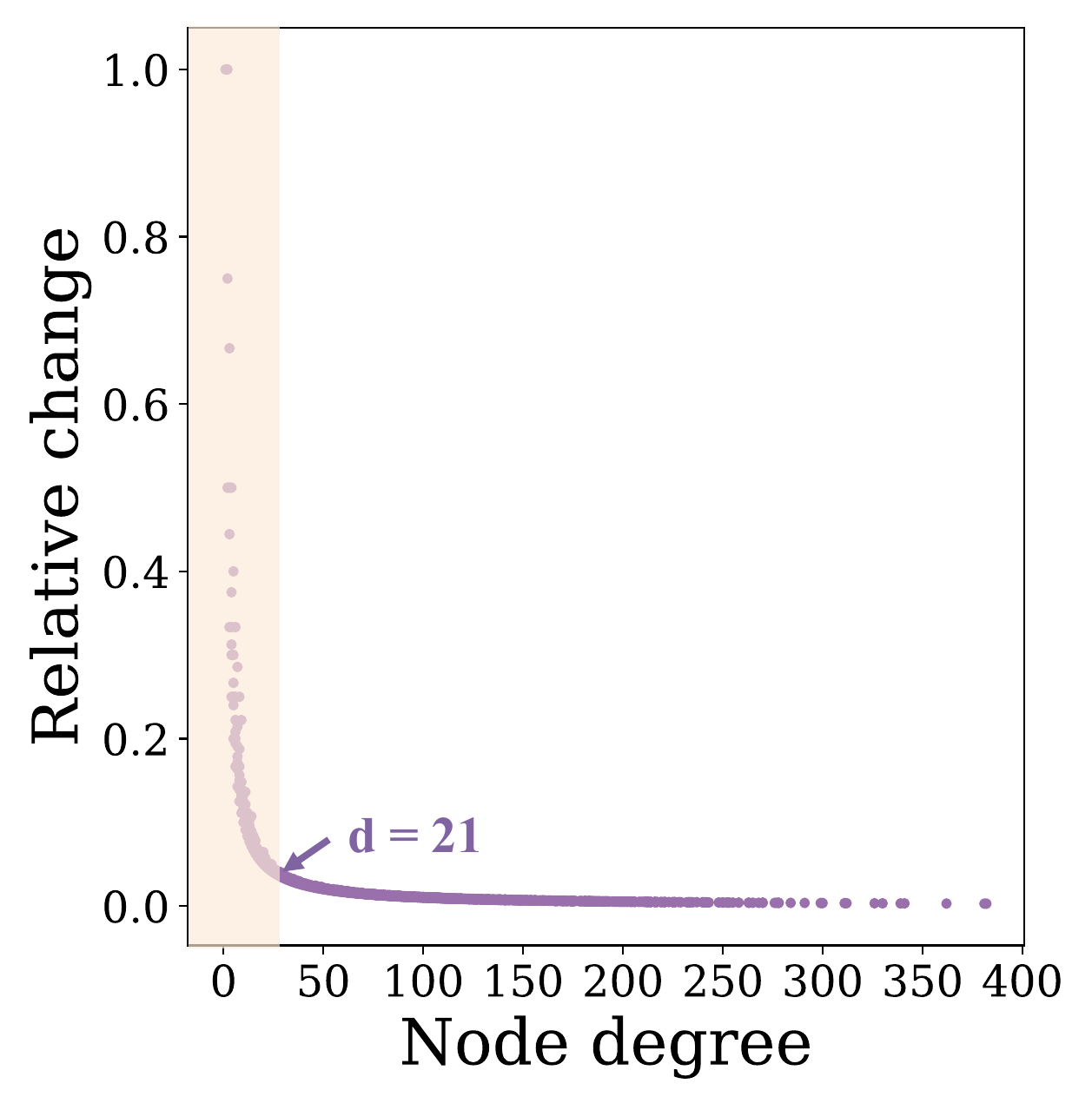}
    \vspace{-0.7em}
    \caption{Ogbl-Collab}
    \label{fig:collab_change}
    \end{subfigure}
    ~ 
    \begin{subfigure}{0.25\linewidth}
    \centering
    \includegraphics[width=0.85\columnwidth]{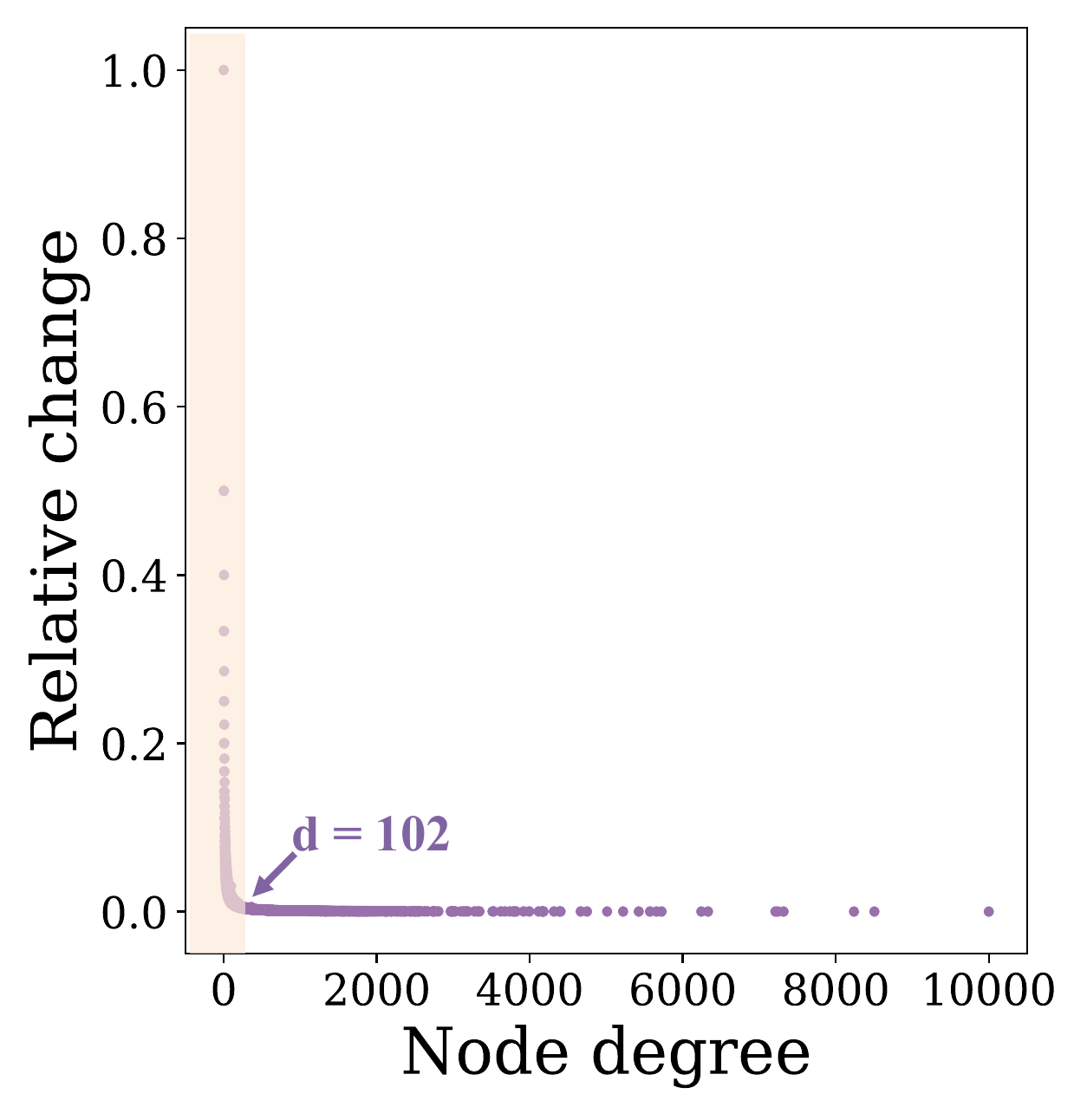}
    \vspace{-0.7em}
    \caption{Ogbl-Citation2}
    \label{fig:citation2_change}
    \end{subfigure}
    ~
    \begin{subfigure}{0.25\linewidth}
    \centering~\includegraphics[width=0.85\columnwidth]{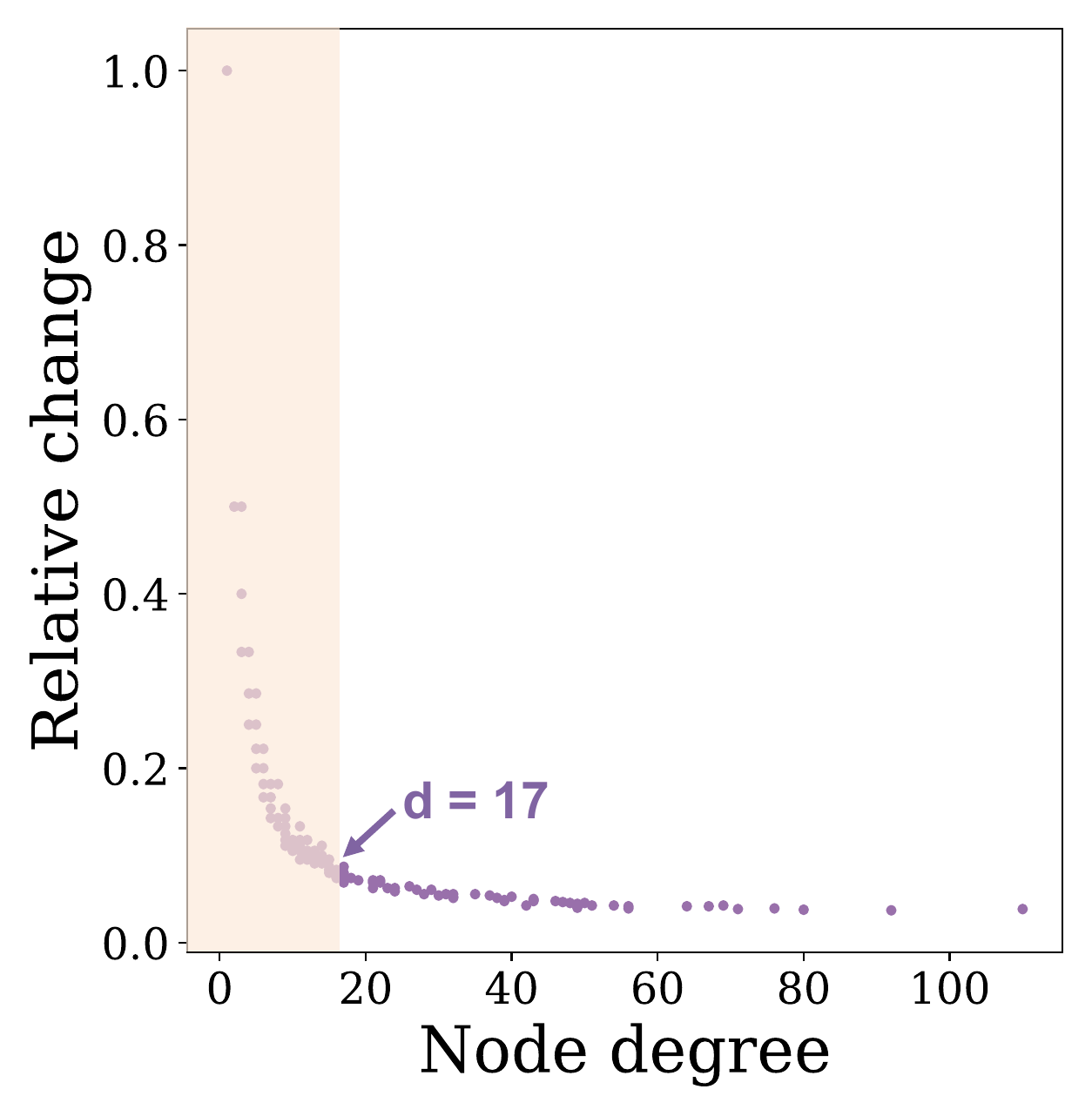}
    \vspace{-0.7em}
    \caption{USAir}
    \label{fig:usair_change}
    \end{subfigure}
    ~
    \begin{subfigure}{0.25\linewidth}
    \centering~\includegraphics[width=0.85\columnwidth]{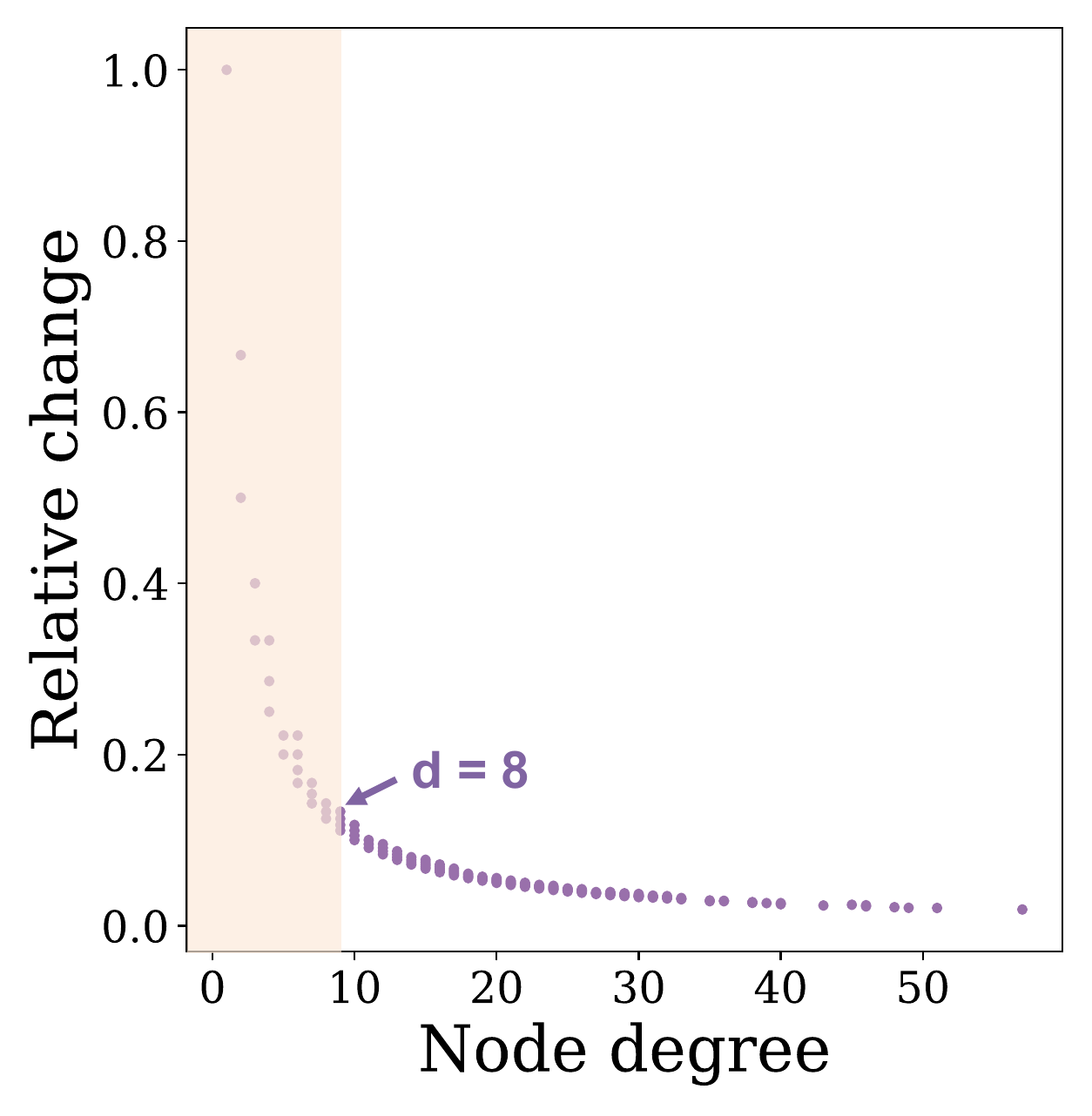}
    \vspace{-0.7em}
    \caption{E-commerce}
    \label{fig:esci_change}
    \end{subfigure}

\vspace{-1.2em}
\caption{{Average degree change for nodes when excluding training target links.} 
The Y-axis corresponds to the relative change in degree before and after excluding all of the train target links in each mini-batch. 
Lower-degree nodes have higher relative degree change; for nodes with degree less than 5, the relative degree change is as high as 100$\%$.}
\vspace{-0.9em}
\label{fig:relative_degree_change}
\end{figure*}

\vspace{-0.2em}
In this section, we present \method, the first framework that systematically resolves the issues arising from the presence of target links in the message-passing graph for link prediction. We propose separate solutions that are tailored to training and inference time. %

\vspace{-0.5em}
\subsection{Training-time Solution: Exclude Target Links Incident to Low-degree Nodes} %
\label{sec:low-degree}

As discussed in Sec. ~\ref{sec:train-pitfalls}, the practice of including train target links $T_{\text{Tr}}$ as message-passing edges causes overfitting and distribution shift (I1-I2). One straightforward solution is to exclude all train target edges during training. However, this poses several challenges for both mini-batch and full-batch settings:
\vspace{-0.2em}
\begin{itemize}
    \item First,  for mini-batch training, an excluded link could be a message-passing edge of another target edge. For example, in Fig. \ref{fig:2-hop graph}, $e_{14}$ is both a target edge and a message-passing edge for node 4 and node 1. The existence of $e_{14}$ affects the message-passing graphs, and, in turn, the learning of target edges $e_{46}$, $e_{01}$ and $e_{12}$. Excluding all the target edges causes significant corruption of the graph structure. In an extreme case, some nodes become isolated nodes, %
    such as node 2 in Fig.~\ref{fig:2-hop graph}. As a result, GNN models may fail to learn good representations when all target edges are excluded.
    \item Second, in full-batch training, if all edges are used as training target edges, then excluding all edges will result in a graph with only nodes and no edges, which is impractical. If only a portion of edges are used as training edges, the graph structure corruption caused by excluding all target edges still applies to full-batch settings. In full-batch training, it requires iterating over all edges in the graph to remove the target edges per training step, which is especially time-consuming. In practice, full batch training for link prediction on massive graphs is rare, since it is inefficient in terms of time and space complexity. As a result, we only consider mini-batch training in our proposed  framework, \method. 
    \item Third, although setting the batch size to 1 can solve the structure corruption, the mini-batch message-passing graph would become too small, causing inefficiency and instability for GNN training.
\end{itemize}

The question then becomes: \textit{How can we achieve the best trade-off between avoiding issues (I1, I2) caused by the presence of train target links and preserving the graph structure in mini-batch training as much as possible?}
The key insight to tackle this problem lies in identifying which nodes are mostly affected by issues (I1, I2), and only excluding target links incident to those nodes.  
At a high level, we show theoretically and empirically that low-degree nodes are impacted most by the inclusion of target link edges, as it causes more significant relative degree changes for them compared to other nodes.  Excluding the target links incident to low-degree nodes achieves the best trade-off: since they have few neighbors, there is generally a small probability that the excluded target links that are incident to them are message-passing edges of another node in the mini-batch training. 
Next, we provide a theoretical and quantitative analysis to show that low-degree nodes are affected most by the issues, and target links in $T_{\text{low}}$ should be excluded during training.

\vspace{0.1cm}
\noindent \textbf{Theoretical Analysis}.
We begin by explaining from a theoretical perspective why primarily low-degree nodes suffer from the issues caused by the inclusion of train targets compared to high-degree nodes. %
Intuitively, we compare the change in influence that a random node $v_k$ has on a high-degree node $v_h$ and a low-degree node $v_l$ before and after excluding an edge incident to $v_h$ and $v_l$, respectively. 
We leverage the notion of influence/effect functions in statistics~\cite{xu2018representation,tang2020investigating} to 
measure the relative influence of a node on another node through a specific train edge.

\vspace{-0.5em}
\begin{theorem}

Let $v_h$ and $v_l$ be two nodes in a graph with degrees $d_h > d_l$, and node $v_k$ be an arbitrary node in the graph. Assume that ReLU is the activation function, the $\Lambda$-layer GNN is untrained, and all random walk paths have a return probability of 0. We denote the effect of node $v_k$ on node $v_h$ after $\Lambda$-th layer GNN as $\pdv{x_h^\Lambda}{x_k}$ where $x_h, x_k$ are $n-$dimensional vectors indicating the embeddings for nodes $v_h, v_k$, respectively. Further we denote that effect of node $v_k$ on node $v_h$ after removing an incident edge to node $v_h$ as $\pdv{\tilde{x_h}^\Lambda}{x_k}$. We define the change in effect of $v_k$ on $v_h$ before and after removing an incident edge to $v_h$  as distance function $D(k, h) = 1- \mathbb{E}(\pdv{\tilde{x}_{h,s}^\Lambda}{x_{k,t}} / \pdv{x_{h,s}^\Lambda}{x_{k,t}})$ for any entry $1 \leq s,t \leq n$ of $x_h$ and $x_k$. Similarly,  we define the change in effect of node $v_k$ on $v_l$ as  $D(k, l) = 1- \mathbb{E}(\pdv{\tilde{x}_{l,s}^\Lambda}{x_{k,t}} / \pdv{x_{l,s}^\Lambda}{x_{k,t}})$ for any entry $1 \leq s,t \leq n$ of $x_l$ and $x_k$.  Then, $D(k, h) < D(k, l)$.  

\label{theorem}
\end{theorem}

Theorem \ref{theorem} states that the change in influence of a random node $v_k$ on another node $v$, caused by excluding a target link is higher on the low degree nodes $v_l$.
This suggests that low-degree nodes benefit more by excluding target edges: when all target edges are present, low-degree nodes are more vulnerable to the issues brought by the inclusion of target edges. 
This statement holds for any GNN model relying on message passing.  We provide detailed proofs here~\footnote{https://arxiv.org/abs/2306.00899}.

\noindent \textbf{Quantitative Analysis: Average degree change}.
We further support our claim that low-degree nodes are affected more by providing a quantitative analysis on the relative degree changes before and after excluding the train target links. We analyze four datasets of various sparsity levels, as shown in Tab. ~\ref{tab:dataset}. For each dataset, we sort its nodes by their degrees and report the average degree change before and after excluding the train targets for each mini-batch epoch. As shown in Fig. ~\ref{fig:relative_degree_change}, for low-degree nodes, the relative change is near 100 $\%$, while for high-degree nodes it is less than 10$\%$.

\noindent \textbf{Proposed Solution: Exclude $T_{\text{low}}$}. To achieve the best trade-off between avoiding issues (I1)-(I2) and minimizing the corruption of the graph structure in mini-batch training, \method excludes $T_{\text{low}}$, the train target edges where at least one incident node has degree lower than a degree threshold $\delta$.  
Implementation-wise, to ensure the scalability and usability of our proposed solution in large-scale, real-world applications, we implemented it as a subclass of DGL's edge sampler, which is comparable to DGL's original edge sampler and can be readily combined with other DGL functions. %

\vspace{-0.5em}
\subsection{Test-time Best Practice: Exclude All Test Target Links}
As we have discussed, including the test target links in the test message-passing graph causes test data leakage. This may occur 
inadvertently---for example, when adapting GNNs designed for node classification tasks for the link prediction task by simply stacking a decoder function---or  
when test target links are explicitly added into the graph to ensure that there is no distribution shift issue. 
We argue that under no circumstance should the test edges be used as message-passing edges. This would ensure more accurate estimation of GNN's predictive performance.

\begin{algorithm}[t!]
\caption{\method: Leakage Check($G$)}
\small 
\label{alg:leakage}
\begin{algorithmic}[1]
\State \textbf{Input:} An input graph $\textbf{G}$, edge splits $\textbf{S}$, an argument $\textbf{K}$ if validation target edges are used as inference inputs, $\textbf{K}= \{T,F\}$
\State \textbf{Output:} The desired inference graph  $\textbf{G}_{\text{infer}}$

\Statex \textbf{\sf \textcolor{gray}{// STEP 1. Check if the input graph contains validation and test edges}}
\State $C_{\text{valid}} = \text{Check Existence}(\textbf{G}, \textbf{S}_{\text{valid}}) $
\State $C_{\text{test}} = \text{Check Existence}(\textbf{G}, \textbf{S}_{\text{test}}) $ 
\Statex \textbf{\sf \textcolor{gray}{// STEP 2. Delete test 
 and validation edges according to user needs}}
\If{$C_{\text{test}}$ is True} 
    \State  $\textbf{G}_{\text{infer}} = \text{RemoveEdge}(\textbf{G},\textbf{S}_{\text{test}})$
\Else
    \State  $\textbf{G}_{\text{infer}} =\textbf{G} $ 
\EndIf
\Statex \textbf{\sf \textcolor{gray}{// If Validation edges exist in the inference graph and it is not desired}}
\If{$C_{\text{valid}}$ is True and $\textbf{K}$ is False}  
    \State  $\textbf{G}_{\text{infer}} = \text{RemoveEdge}(\textbf{G}_{\text{infer}},\textbf{S}_{\text{valid}})$
\EndIf

\State \Return{$\textbf{G}_{\text{infer}}$}
\end{algorithmic}
\vspace{-0.4em}
\end{algorithm}

\vspace{0.1cm}
\noindent \textbf{Proposed Solution.}  \method excludes all the test target links from the test message-passing graph. 
Moreover, it supports automatically checking for data leakage in user-specified data splits and rectifying the issues as needed (Alg. ~\ref{alg:leakage}). 
In prior work~\cite{hu2020open}, validation edges are sometimes used in the message-passing graphs to obtain more information, especially for data that is split into training/validation/test sets according to time. Including the validation target edges is typically not seen as data leakage. The decision of whether or not to use the validation edges as message-passing edges depends on the application of interest. \method requires the user to deliberately define this design choice, and generates the inference graph that complies with the user requirements.

\section{Experiments}
\label{sec:experiments}

Through our extensive empirical analysis, we aim to address the following research questions:
\begin{itemize}
    \item \textbf{RQ1}: How well does \method address issues (I1) and (I2) on commonly-used graph benchmarks, which are dense? 
    \item \textbf{RQ2}: How well does \method perform on sparse graphs with very skewed degree distributions?   
    \item \textbf{RQ3}: How well does \method address issues (I1)-(I2) for edges incident to low-degree nodes on popular benchmarks?
    \item \textbf{RQ4}: At test time, how much is the performance of GNN models overestimated due to implicit data leakage (I3)?
\end{itemize}

Before presenting our results, we describe the experiment setup.

\vspace{0.1cm}
\noindent\textbf{Data}.
We evaluate our framework on four real-world datasets on the link prediction task and give their statistics in Tab.~\ref{tab:dataset}. Ogbl-Collab and Ogbl-Citation2~\cite{hu2020open} are  author collaboration and citation networks. 
USAir~\cite{ribeiro2017struc2vec} is a network of US airlines.
We note that these datasets are relatively dense, with average node degree of 8-20.
In real-world applications, the observed data is typically incomplete and sparse, with skewed degree distributions and many low-degree nodes.
For this reason, we also consider E-commerce~\cite{reddy2022shopping}, a sparse real-world dataset of queries and related products that are exact matches in Amazon Search. %

\begin{table}[t]
\centering
\caption{Dataset statistics based on the training splits.  %
}
\vspace{-1.2em}
\label{tab:dataset}
\resizebox{\linewidth}{!}{
\begin{tabular}{lrrrr} 
\toprule
Dataset & \# Nodes & \# Edges & Node deg. & Attr. dim. \\ 
\midrule
ogbl-collab~\cite{hu2020open} & 235,868 & 2,358,104 & 8.20 & 128 \\
ogbl-citation2~\cite{hu2020open} & 2,927,963 & 30,387,995 & 20.73 & 128 \\
USAir~\cite{ribeiro2017struc2vec}& 332 & 3,402 & 10.25 & 332 \\
E-commerce ~\cite{reddy2022shopping} & 346,439 & 238,818 & 1.38 & 768 \\ 
\bottomrule
\end{tabular}
}
\vspace{-1.5em}
\end{table}

\begin{table}[t!]
\centering
\caption{{RQ1-Training Issues: Results on dense graphs.} Test performance of different training frameworks across GNNs and datasets.
\method has the best overall performance (lowest rank) across all datasets. 
*OOM = out of GPU memory.}
\vspace{-1.2em}
\label{tab:excluding_results}
\resizebox{\columnwidth}{!}
{
\begin{tabular}{@{}l cccc@{}}
\toprule
\multirow{2}{*}{\textbf{Model}} & \textbf{ExcludeNone(Tr)} & \textbf{ExcludeAll} & \textbf{ExcludeRandom}  & \textbf{\method}\\
 \cline{2-5}
 & \multicolumn{4}{c}{\textbf{Ogbl-Collab (H@50 $\uparrow$)}} \\  
\midrule
\textbf{SAGE} & 48.57 $\pm$ 0.74 & 45.82 $\pm$ 0.41 & 45.74 $\pm$ 1.33 & \colorbox{Gray}{49.00 $\pm$ 0.65}  \\

\textbf{MB-GCN} & \colorbox{Gray}{43.03 $\pm$ 0.50} & 37.75 $\pm$ 1.42 & 41.43 $\pm$ 2.25 & 39.58 $\pm$ 1.06 \\

\textbf{GATv2} & 45.61 $\pm$ 0.85  & 45.71 $\pm$ 0.87 & \colorbox{Gray}{45.87 $\pm$ 0.64} & 45.46 $\pm$ 0.19 \\

\textbf{SEAL} & 61.27 $\pm$ 0.28 & 64.11 $\pm$ 0.30 & 64.40 $\pm$ 0.57 & \colorbox{Gray}{64.57 $\pm$ 0.30}  \\

 \midrule
  & \multicolumn{4}{c}{\textbf{Ogbl-Citation2 (MRR $\uparrow$)}} \\ 
  \midrule
\textbf{SAGE} & 82.06 $\pm$ 0.06 & 81.47 $\pm$ 0.17 & 82.06 $\pm$ 0.13 & \colorbox{Gray}{82.18 $\pm$ 0.18}  \\

\textbf{MB-GCN} & 79.70 $\pm$ 0.25 & 79.06 $\pm$ 0.30 & \colorbox{Gray}{80.39 ± 0.15} & 79.88 $\pm$ 0.14\\

\textbf{GATv2} & OOM  & OOM & OOM & OOM \\

\textbf{SEAL} & 86.75 $\pm$ 0.20 & 86.74 $\pm$ 0.23 & 86.61 ± 0.39 & \colorbox{Gray}{86.93 $\pm$ 0.55}  \\

 \midrule
  & \multicolumn{4}{c}{\textbf{USAir (AUC $\uparrow$)}} \\
  \midrule
\textbf{SAGE} & 95.97 $\pm$ 0.17 & 95.71 $\pm$ 0.12 & \colorbox{Gray}{96.42 $\pm$ 0.18} & 96.19 $\pm$ 0.53  \\

\textbf{MB-GCN} & 94.00 $\pm$ 0.14 & 94.09 $\pm$ 0.11 & 93.98 ± 0.06 & \colorbox{Gray}{94.28 $\pm$ 0.15}\\

\textbf{GATv2} & 95.05 $\pm$ 0.66  & 95.66  $\pm$ 0.24 & 95.80 $\pm$ 0.24 & \colorbox{Gray}{95.87 $\pm$ 0.46} \\

\textbf{SEAL} & 95.36 $\pm$ 0.24 & 95.94 $\pm$ 0.04 & 95.76 $\pm$ 0.24 & \colorbox{Gray}{96.39 $\pm$ 0.09}  \\

\midrule
\textbf{Rank $\downarrow$} & 2.81 & 3.09 & 2.45 & \colorbox{Gray}{1.64} \\
\bottomrule
\end{tabular}
}
\vspace{-1.5em}
\end{table}

\begin{table*}[t!]
\centering
\caption{{RQ2-Training Issues: Results on the sparse E-commerce dataset.} 
\method achieves consistently better performance than the baseline across metrics and models. For SAGE and GATv2, \method is up to 15× more accurate. }
\label{tab:esci}
\vspace{-1.4em}
{
\begin{tabular}{@{}l@{\hskip15pt}c@{\hskip15pt}c c@{\hskip15pt} c@{\hskip15pt}c c@{\hskip15pt} c@{\hskip15pt}c@{}}
\toprule
 & \multicolumn{2}{c}{\textbf{SAGE}} && \multicolumn{2}{c}{\textbf{MB-GCN}} && \multicolumn{2}{c}{\textbf{GATv2}} \\ \cline{2-3} \cline{5-6} \cline{8-9}
 \textbf{Metrics} & \textbf{ExcludeNone(Tr)} & \textbf{\method} && \textbf{ExcludeNone(Tr)} & \textbf{\method} && \textbf{ExcludeNone(Tr)} & \textbf{\method}\\
 \midrule
 \textbf{MRR $\uparrow$} &  4.40 $\pm$ 0.31 &  \colorbox{Gray}{65.85 $\pm$ 0.31} &&  17.07 $\pm$ 7.38 & \colorbox{Gray}{69.67 $\pm$ 0.52} &&  5.98 $\pm$ 0.56 & \colorbox{Gray}{69.44 $\pm$ 0.55} \\
 \textbf{H@10 $\uparrow$} &  6.55 $\pm$ 0.37 &  \colorbox{Gray}{89.67 $\pm$ 0.19} &&   28.35 $\pm$ 7.47 & \colorbox{Gray}{89.79 $\pm$ 0.25} &&  9.64 $\pm$ 1.10 & \colorbox{Gray}{90.52 $\pm$ 0.26} \\
 \textbf{H@1 $\uparrow$} &  3.04 $\pm$ 0.31 &  \colorbox{Gray}{52.84 $\pm$ 0.46} &&   10.83 $\pm$ 5.21 & \colorbox{Gray}{57.63 $\pm$ 0.57} &&  3.94 $\pm$ 0.81 & \colorbox{Gray}{57.11 $\pm$ 1.03} \\
\bottomrule
\end{tabular}
}
\vspace{-1.7em}
\end{table*}

\vspace{0.1cm}
\noindent \textbf{Metrics}.
Following prior works, we use Mean Reciprocal Rank (MRR) on Ogbl-Citation2 and Hits@50 on Ogbl-Collab~\cite{hu2020open}. Area Under the Curve (AUC) is used for USAir ~\cite{zhang2018link}. For E-commerce, we choose to report MRR, Hits@10, and Hits@1, the three most commonly-used evaluation metrics for link prediction~\cite{hu2020open,zhang2021labeling,vashishth2019composition}. 
For all evaluation metrics, the higher the value is, the better.

\vspace{0.1cm}
\noindent \textbf{GNN models}.
We select four GNN models to validate our proposed solutions. SAGE~\cite{hamilton2017inductive}, MB-GCN~\cite{kipf2016semi} and GATv2~\cite{brody2021attentive} are GAE-based models. 
MB-GCN~\cite{kipf2016semi} is a mini-batch GCN model and at each iteration, only a portion of the entire graph is seen. 
SEAL~\cite{zhang2018link} is a subgraph-based model that extracts an enclosing subgraph for each target edge and predicts the link likelihood based on the subgraph's embeddings. All GNNs are implemented in DGL~\cite{paszke2017automatic, dgl}. We conduct a hyperparameter tuning and choose the best.

\vspace{0.1cm}
\noindent \textbf{Baselines}.
For training-time issues (I1, I2), we use ExcludeNone(Tr), ExcludeAll and ExcludeRandom as our baselines.
ExcludeNone(Tr) does not exclude any training target links, while ExcludeAll excludes all target links $T_{\text{Tr}}$. Note that ExcludeAll on SEAL is essentially FakeEdge, which excludes all target edges on subgraph-based models. ExcludeRandom randomly excludes target edges during training, and the proportion of excluded targets is the same as our \method.
For test-time issues (I3), our baseline is ExcludeNone(Tst), which uses the test target links in the inference graph. This approach corresponds to the case where data leakage occurs, which should always be avoided in real-world applications. %

\vspace{0.1cm}
\noindent \textbf{\method Variants}. 
At training time, we consider two variants of \method that differ in the degree threshold $\delta$ that they use to exclude target links $T_{\text{low}}$ for all datasets, and report the best-performing one: $\delta=10$ or $\delta=20$, which corresponds to the average degrees of the dense datasets we used. 
For the E-Commerce dataset, since 99.5\% edges are incident to nodes with degree less than 5, \method excludes almost all target edges and achieves similar impact as ExcludeAll and ExcludeRandom.
At test time, we consider two variants for \method:  %
ExcludeValTst excludes both validation and test target edges from the test graph, while ExcludeTst only excludes the test target links. 
As shown in Alg. ~\ref{alg:leakage}, whether to use ExcludeValTst or ExcludeTst depends on the user's input.
\subsection{RQ1-Training Issues: Results on Dense Data}
\noindent \textbf{Setup}. To evaluate \method's ability to address training issues (I1) and (I2) on dense graphs, we report the link prediction performance of four GNN models on three popular dense datasets over three trials. We report the recommended metrics for each dataset. For Ogbl-Collab and Ogbl-Citation2, we generate one negative per target edge during training and use the recommended negatives during evaluation.  For USAir, we also generate one negative per target edge during training, while during evaluation, we treat all edges that do not appear in the train,test,validation as negative edges. 
In addition to the performance for each setting, we also report the average rank of our baselines  and proposed framework \method.  
Our results are summarized in Tab. ~\ref{tab:excluding_results}.

\noindent \textbf{Results}. \method achieves the best performance (lowest rank) across datasets and models. On Ogbl-Citation2 and USAir, our method almost achieves the best results across different types of models. This indicates that \method successfully addresses the train issues (I1, I2) while also avoiding significant corruption of the structure in the mini-batch graphs.
Although the original implementation of SEAL uses ExcludeAll, we find that replacing that strategy with \method further helps improve SEAL's performance. %

Moreover, comparing \method with ExcludeRandom, we can see that \method consistently gives better performance. This experimentally verifies Theorem \ref{theorem} and show that specifically excluding the edges incident to low-degree nodes can benefit more.

We also observe that ExcludeAll typically results in slightly lower performance compared to ExcludeNone(Tr). As discussed in Sec. ~\ref{sec:low-degree}, this is mainly because excluding all target edges in one mini-batch causes a significant change on graph structure and even isolates some nodes. Thus, GNNs will not learn good node representations. 
\vspace{-0.5em}
\begin{observation}
\textbf{\emph{(1)}} Across all datasets and models, \method achieves the best overall rank compared with ExcludeNone(Tr), ExcludeAll and ExcludeRandom. This indicates that it successfully addresses the issues (I1) and (I2).   
\textbf{\emph{(2)}} In many cases (6/11), ExcludeAll leads to performance degradation because of currupting the structure of mini-batch graphs. 
\end{observation}

\begin{table*}[t!]
\centering
\caption{{RQ3-Training Issues: Results on low-degree nodes.} We report MRR of SAGE on Ogbl-Citation2 on target edges incident to at least one low-degree nodes ($min(d_i,d_j)$) or only low-degree nodes ($max(d_i,d_j)$).  \method achieves the best performance. }
\vspace{-1em}
\label{tab:low_degree_results}
\resizebox{\textwidth}{!}
 {
\begin{tabular}{llcccccc}
\toprule
& \textbf{Exclusion} & \textbf{$max(d_i, d_j) < 10$} & \textbf{$max(d_i, d_j) < 5$} & 
\textbf{$min(d_i, d_j) < 10$} &
\textbf{$min(d_i, d_j) < 5$} &
\textbf{$min(d_i, d_j) = 2$} &
\textbf{$min(d_i, d_j) = 1$} \\
  \midrule
 \multirow{3}{*}{\textbf{MRR} $\uparrow$}& \textbf{ExcludeNone(Tr)} & $73.11 \pm 0.25$ & $62.15 \pm 0.84$ & $78.78 \pm 0.12$ & $69.54 \pm 0.37$ & $47.02 \pm 0.56$ & $27.54 \pm 0.88$\\

& \textbf{ExcludeAll} & $77.45 \pm 0.41$ & $75.39 \pm 1.42$ & $79.17 \pm 0.12$ & $73.86 \pm 0.33$ & $60.05 \pm 1.11$ & $48.60 \pm 1.11$\\

& \textbf{ExcludeRandom} & $76.11 \pm 0.12$ & $70.79 \pm 0.53$ & \colorbox{Gray}{$79.41 \pm 0.06$} & $72.31 \pm 0.04$ & $55.21 \pm 0.06$ & $41.48 \pm 0.42$\\

& \textbf{\method} & \colorbox{Gray}{$78.08 \pm 0.06$} & \colorbox{Gray}{$76.23 \pm 0.56$} & $79.30 \pm 0.18$ & \colorbox{Gray}{$73.87 \pm 0.18$} & \colorbox{Gray}{$61.48 \pm 0.51$} & \colorbox{Gray}{$51.47 \pm 2.51$}\\
\bottomrule

\end{tabular}
}
\vspace{-1.2em}
\end{table*}

\begin{table}[t!]
\centering
\caption{{RQ4-Test Issue: Leakage quantification.} We report the test results of four GNNs over three datasets. 
Note that ExcludeNone(Tst)'s good performance is due to data leakage; the test edges, never observed in real-world applications, are used during inference. 
Using test target links should be avoided; our framework, \method, can automatically check and/or enforce this. *OOM = out of GPU memory. 
}
\vspace{-1.1em}
\label{tab:leakage_results}
\resizebox{\columnwidth}{!}
{
\begin{tabular}{@{}L{0.22\columnwidth}D{0.25\columnwidth}O{0.22\columnwidth}R{0.3\columnwidth}@{}}
\toprule
\multirow{2}{*}{\textbf{Models}} & \multicolumn{2}{c}{\textbf{\method}} & \cellcolor{white} \textbf{Baseline}\\ \cmidrule{2-3}
& \textbf{ExcludeValTst} & \textbf{ExcludeTst} & \textbf{ExcludeNone(Tst)}\\
\midrule
 & \multicolumn{3}{c}{\textbf{Ogbl-Collab (H@50 $\uparrow$)}} \\  

  \cmidrule{2-4}

\textbf{SAGE} & 48.57 $\pm$ 0.74 & 57.61 $\pm$ 0.88 & \textcolor{ggray}{83.82 $\pm$ 0.59} \\

\textbf{MB-GCN} & 43.03 $\pm$ 0.50 & 50.53 $\pm$ 1.10 & \textcolor{ggray}{75.41 $\pm$ 0.43}\\

\textbf{GATv2} & 45.61 $\pm$ 0.85 & 54.94 $\pm$ 0.19 & \textcolor{ggray}{84.16 $\pm$ 2.62} \\

\textbf{SEAL} & 57.50 $\pm$ 0.31 & 55.16 $\pm$ 1.94 & \textcolor{ggray}{99.91 $\pm$ 0.05} \\

 \cmidrule{2-4}
 & \multicolumn{3}{c}{\textbf{Ogbl-Citation2 (MRR $\uparrow$)}} \\  

 \cmidrule{2-4}

\textbf{SAGE} & 82.06 $\pm$ 0.06 & 82.28 $\pm$ 0.11 & \textcolor{ggray}{89.22 $\pm$ 0.10} \\

\textbf{MB-GCN} & 79.70 $\pm$ 0.25 & 81.25 $\pm$ 0.22 & \textcolor{ggray}{88.32 $\pm$ 0.14} \\

\textbf{GATv2} & OOM & OOM & \textcolor{ggray}{OOM} \\

\textbf{SEAL} & 86.75 $\pm$ 0.20 & 87.01 $\pm$ 0.39 & \textcolor{ggray}{97.14 $\pm$ 0.18} \\
 \cmidrule{2-4}

 & \multicolumn{3}{c}{\textbf{USAir (AUC $\uparrow$)}} \\  

 \cmidrule{2-4}

\textbf{SAGE} & 95.97 $\pm$ 0.17 & 95.51 $\pm$ 0.53 & \textcolor{ggray}{99.15 $\pm$ 0.59} \\

\textbf{MB-GCN} & 94.00 $\pm$ 0.14 & 94.11 $\pm$ 0.13 & \textcolor{ggray}{98.66 $\pm$ 0.22} \\

\textbf{GATv2} & 95.05 $\pm$ 0.66 & 94.07 $\pm$ 0.21 & \textcolor{ggray}{98.96 $\pm$ 0.11} \\

\textbf{SEAL} & 95.36 $\pm$ 0.24 & 95.10 $\pm$ 0.76 & \textcolor{ggray}{97.20 $\pm$ 0.78} \\

\midrule
\textbf{No Leakage?} & \textcolor{green}{\cmark} & \textcolor{green}{\cmark} & \textcolor{red}{\xmark}  \\
\textbf{Deployment} & \textcolor{green}{\cmark} & \textcolor{green}{\cmark} & \textcolor{red}{\xmark}  \\
\bottomrule
\end{tabular}
}
\vspace{-1.5em}
\end{table}

\vspace{-1em}
\subsection{RQ2-Training Issues: Results on Sparse Data}

\noindent \textbf{Setup}.  In the real-world E-commerce dataset, the graph is incomplete, sparse and full of low-degree nodes. Based on our theoretical analysis in Sec.~\ref{sec:low-degree}, the low-degree nodes suffer more from training issues.
To investigate the usefulness of \method in such settings, we repeat the previous experiments.  Note that we do not report the results of ExcludeAll and ExcludeRandom because almost all edge is incident to nodes with degree less than 5, so \method excludes nearly every target edge. Furthermore, due to the high sparsity, we also do not report the results for SEAL since it is impractical to construct subgraphs for each node. The results are shown in Tab. ~\ref{tab:esci}.

\noindent \textbf{Results}. On sparse graphs like E-commerce, \method achieves \textit{14.9 $\times$} better performance. Since many real-world graphs are very sparse (e.g. commonsense knowledge graphs and biochemical graphs have an average degree of 2 ~\cite{malaviya2020commonsense, dwivedI2022long}), \method can improve the performance of GNNs across numerous high-impact settings.

\vspace{-0.5em}
\begin{observation}
\method achieves \textit{14.9$\times$} better performance compared to ExcludeNone across models. This verifies empirically that low-degree nodes suffer more from issues (I1) and (I2), and excluding $T_{\text{low}}$ works well especially for %
datasets with many low-degree nodes. 
\end{observation}

\vspace{-0.8em}
\subsection{RQ3-Training Issues: Results on Low-degree Nodes}
\label{sec:rq3}
\noindent \textbf{Setup.} 
To quantify how much low-degree nodes in dense datasets suffer from issues (I1) and (I2), we explore the predictive performance for edges adjacent to low-degree nodes. We report the performance of two different edge types: (1) edges that are incident to at least one low-degree node, i.e., $min(d_i, d_j) < \delta$ and; 
(2) edges that are only incident to low-degree nodes, i.e.,  $max(d_i, d_j) < \delta$. We report results on Ogbl-Citation2 for SAGE, and compare \method against three baselines. 
Results are shown in Tab. \ref{tab:low_degree_results}.

\noindent \textbf{Results}.  For edges that are incident to low-degree nodes, ExcludeAll, ExcludeRandom and \method achieve significantly better performance than ExcludeNone(Tr). This corresponds to our theoretical analysis in Sec.~\ref{sec:low-degree} that highlights low-degree nodes are harmed by training issues (I1) and (I2) more, and excluding train target edges is more beneficial for low-degree nodes. Specifically, comparing ExcludeNone(Tr), ExcludeAll and ExcludeRandom, \method achieves better performance on various types of edges that are incident to low-degree nodes. This indicates that \method is better at maintaining the graph structure in mini-batch training. 
\vspace{-0.3em}
\begin{observation}
Better performance on edges adjacent to low-degree nodes in dense graphs indicates that \method successfully resolves (I1, I2) on low-degree nodes. %
\end{observation}

\subsection{RQ4-Test Issues: Leakage Quantification}
\noindent \textbf{Setup}. Beyond the training issues (I1, I2), we aim to quantify the performance gap introduced by the data leakage at test time (I3). To achieve this, we report results on excluding different types of edges from the inference graph (validation, test edges).
Although we are not evaluating in a deployed system, by excluding different types of edges, we are mimicking what would happen in a real application. All GNNs are trained using train edges only. ExcludeValTst excludes all validation and test target links during inference, and ExcludeTst only excludes validation edges. Both ExcludeValTst and ExcludeTst are variants of \method. ExcludeNone(Tst) keeps all validation and test target links during testing, resulting in data leakage (I3) and should be avoided in practice.
The results are shown in Tab. ~\ref{tab:leakage_results}.

\noindent \textbf{Results}.
When validation target links are used as message-passing edges in inference graphs, we observe a slight performance boost, which matches findings in prior work~\cite{hu2020open}. 
However, the performance boost due to the inclusion of the test target links is undesired, as it can lead to overestimation of the models' predictive performance. 
In practice, test links cannot be observed and utilized.  Specifically, when test targets are present during inference, SEAL seemingly achieves near-perfect results, which are not indicative of actual performance. \method successfully resolves issue (I3). 
\vspace{-0.5em}
\begin{observation}

Due to data leakage (I3), using test edges causes a fake performance boost across multiple datasets, especially for those with time-based splits (e.g., Ogbl-Collab). %
Increased performance verifies the necessity of \method, which always excludes the test target links from the inference graphs at test time. Since in real applications, future (test) links are  never observed, if the model utilizes information from test target edges, its performance gets overestimated, i.e., a fake performance boost that will not be seen in practice is achieved.  
\end{observation}
\vspace{-0.8em}

\section{Conclusion}
\label{sec:conclusion}
In this work, we focused on the pitfalls in link prediction with GNNs and systematically study the issues that arise from including the target links as message passing edges. We are the first to show (both theoretically and empirically) that low-degree nodes suffer more from these issues.  %
Our proposed framework, \method, strikes the best balance between eliminating the issues from the target links, not significantly corrupting the structure of the mini-batch graphs, and being scalable and easy to use. 
\method can help researchers and practitioners adhere to best practices, which are frequently overlooked even by the widely-used GNN frameworks.

\section*{Acknowledgments}
We thank Yongyi Yang for 
providing constructive feedback on our theoretical proof. This material is based upon work supported by the National Science Foundation under IIS~2212143,  CAREER Grant No.~IIS 1845491, and AWS Cloud Credits for Research. Any opinions, findings, and conclusions or recommendations expressed in this material are those of the author(s) and do not necessarily reflect the views of the National Science Foundation or other funding parties.
 
\balance
\section*{Ethical Discussion}

In our work, the datasets we used are publicly available; we did not collect or release new datasets. During data pre-processing, we strictly followed ethical principles and did not attempt to infer sensitive attributes. As discussed in Sec.~\ref{sec:rq3}, our approach is able to improve performance for edges adjacent to low-degrees nodes, which can be used to mitigate the potential bias of current GNNs on marginalized nodes (e.g., individuals) that have few connections.
\bibliographystyle{ACM-Reference-Format}
\bibliography{BIB/bibliography}

\clearpage
\appendix
\section{Appendix}
\label{sec:appendix}
\subsection{Experimental Details}
\label{sec:exp-appendix}
\noindent \textbf{E-commerce Dataset Construction}.
The E-commerce dataset is constructed by keeping only links that represent ``exact'' matches between queries and products~\cite{reddy2022shopping}. The queries are randomly divided into train, validate and test sets according to a 70\%/10\%/20\% ratio. 
We use BERT embeddings~\cite{devlin2018bert} as node features. 

\vspace{0.1cm}
\noindent \textbf{Hyperparameter Tuning}.
We conduct extensive hyperparameter tuning using grid search. We search on the learning rates $ = \{$1e-2, 1e-3, 1e-4, 5e-4, 5e-5$\}$ and the number of layers $= \{1,2,3\}$, hidden dimension $=\{128, 256, 512, 1024 \}$. We report the best performing hyperparameters for each setting. We used a Nvidia A40 GPU to train the model and repeat our experiments with three random seeds. Test results are reported on the best-performing validation epoch.
Our result on FakeEdge is lower than reported because (1) we use a different split of USAir due to no public splits available. (2) For FakeEdge, they set the number of hops to 2, hidden channel to 128. We found this to be computationally intensive and cannot be run on larger datasets such as Ogbl-Citation2. We follow the settings from SEAL, and set the number of hops to 1 and the hidden channel to 32 \cite{dong2022fakeedge}.

\begin{figure}[t!]
\centering
    \begin{subfigure}{0.46\linewidth}
    \centering
    \includegraphics[width=\columnwidth]{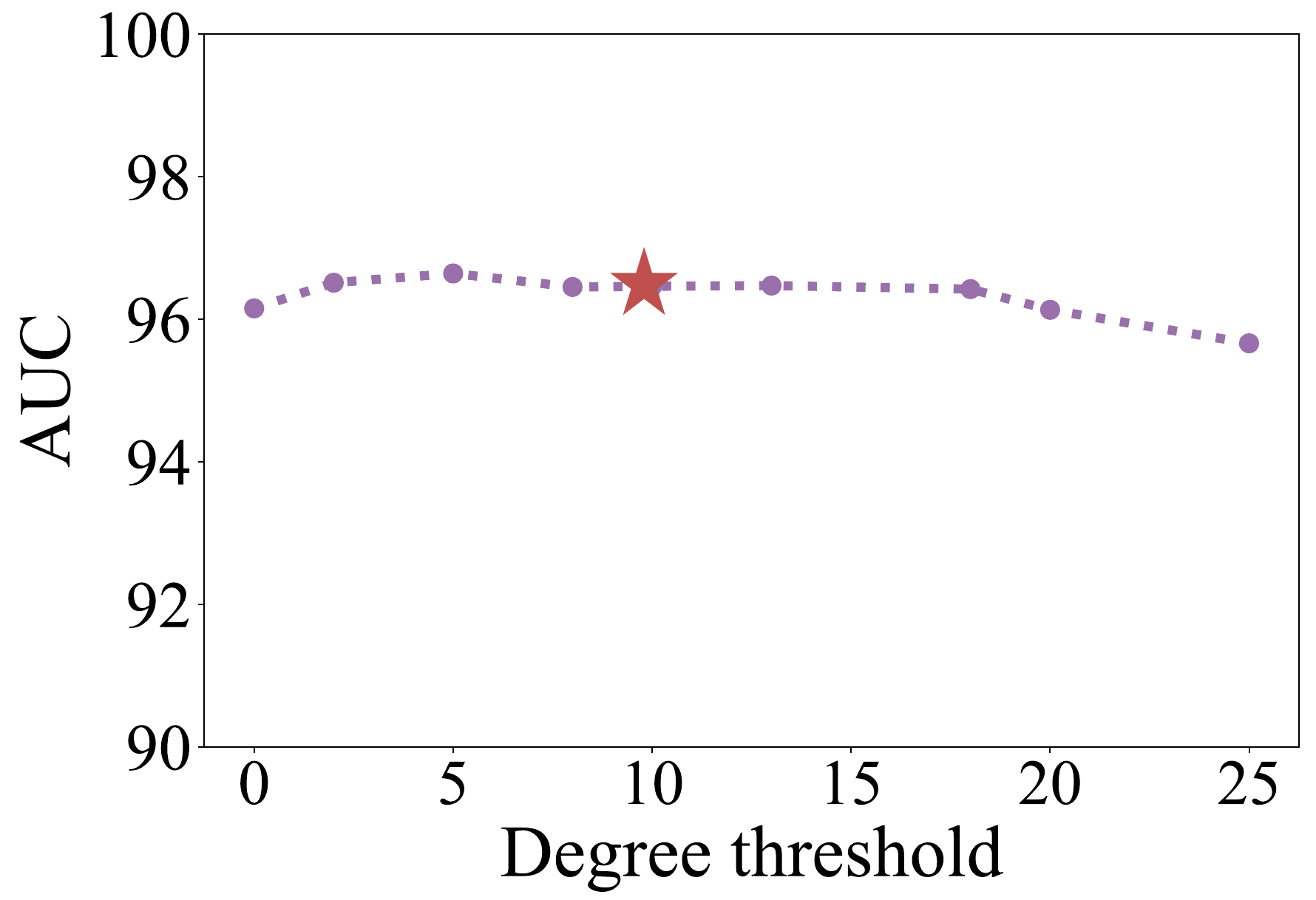}
    \caption{USAir with SAGE model}
    \label{fig:usair_sage}
    \end{subfigure}
    ~ 
    \begin{subfigure}{0.46\linewidth}
    \centering
    \includegraphics[width=\columnwidth]{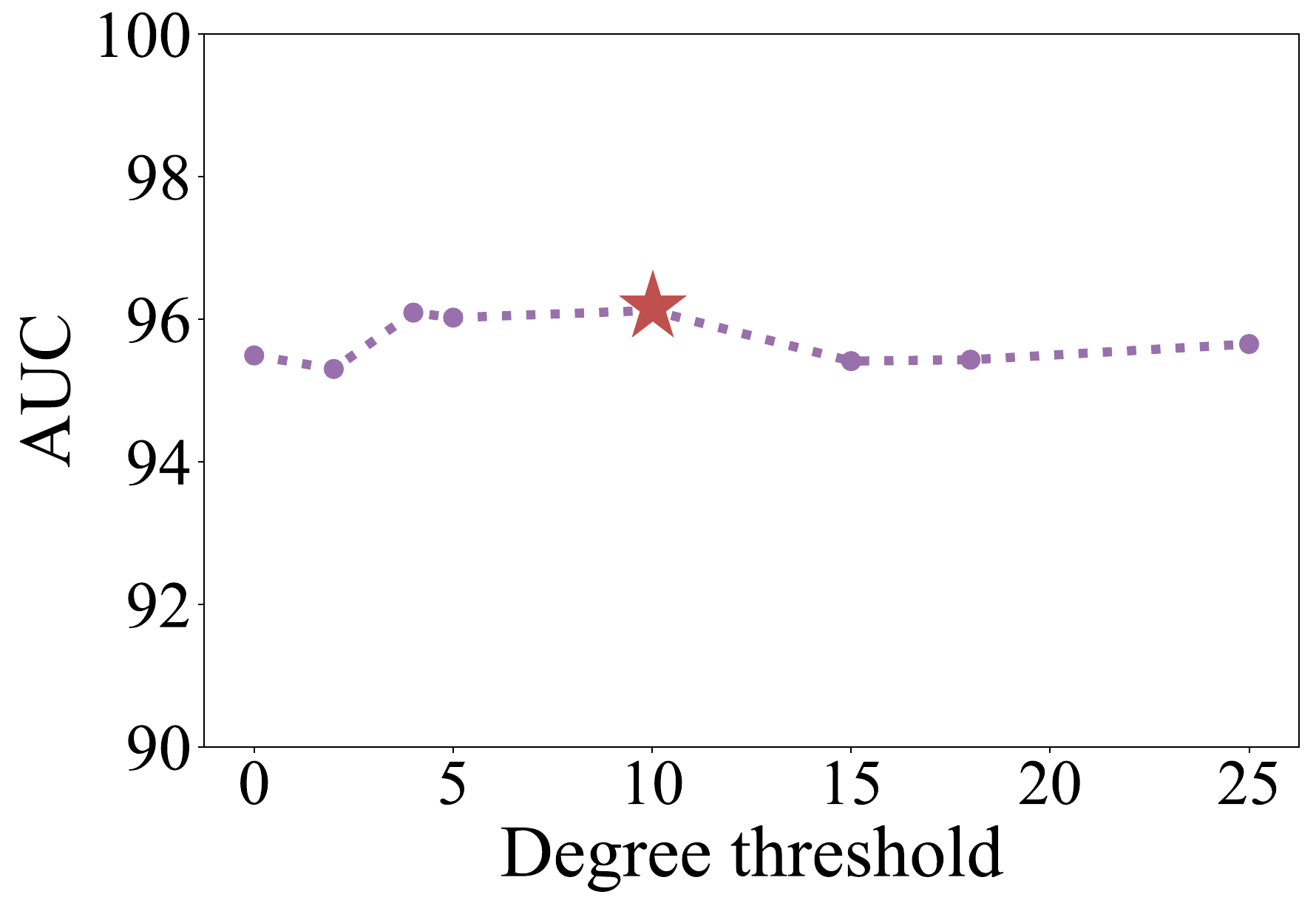}
    \caption{USAir with GATv2 model}
    \label{fig:usair_gat}
    \end{subfigure}

\vspace{-1em}
\caption{\method has robust performance for varying (low) degree thresholds across GNN models. We see a slight `U-shape' effect, which works best when excluding the train target links $T_\text{low}$. The red star indicates the average degree. 
}
\label{fig:degree_with_performance}
\end{figure}

\vspace{0.1cm}
\noindent \textbf{Ablation: Which Degree to Use?} At training time, we only exclude edges adjacent to nodes smaller than a degree threshold $\delta$. One research question that arises is how do we determine the threshold $\delta$? We conduct experiments on USAir with varying $\delta$. The results are shown in Fig. ~\ref{fig:degree_with_performance}. 
As we exclude target edges with a higher degree threshold (exclude more target edges), the performance of the model will first go up and then go down, forming a U-shape curve. This indicates that we need to strike a balance between eliminating the training issues and preserving the structures of the mini-batch graph. A sensitivity check is needed to find the optimal degree threshold. In practice, we found
that choosing $\delta$ as the average degree of dense datasets typically yields good performance.

\vspace{0.1cm}
\noindent \textbf{Time Complexity Analysis}.
The additional time complexity of \method comes from the target edge exclusion part. For each iteration, we need to iterate over the edges in the mini-batch to examine whether they are incident to low-degree nodes. The time complexity of excluding the target edges is $\mathcal{O}(|B|)$, where $|B|$ is the number of edges in the message passing graph. The time complexity of training in the ExcludeNone(Tr), ExcludeAll, ExcludeRandom and \method frameworks is similar since the time of additional edge exclusion is much smaller compared with the model complexity. The difference of number of edges in the message passing graph only makes marginal changes in the training time ~\cite{blakely2021time}.

\vspace{-0.5em}

\subsection{Extended Theoretical Analysis}
We first prove Theorem ~\ref{theorem} on GCN and then extend the proof into general message-passing GNN models. 
\label{sec:extended_proof}

\vspace{-0.5em}
\begin{proof}
We want to prove that when a neighboring edge of a node is removed in order to eliminate the train issues: overfitting (I1) and distribution shift (I2), the changes on high degree nodes is smaller than the change on low degree nodes.  
We first define the overall influence of node $v_k$ on node $v_h$ after $\Lambda$-th layer GCN as $\pdv{x_h^\Lambda}{x_k}$~\cite{xu2018representation,tang2020investigating}.
According to ~\cite{tang2020investigating}, we have that the partial derivative of $x_h$ to $x_k$ for an $\Lambda$-th layer untrained GCN is
\begin{equation}
 \pdv{x_{h,s}^\Lambda}{x_{k,t}} = \sqrt{d_hd_k} \sum_{p=1}^{\Psi} \prod_{\lambda = 0}^{\Lambda}\frac{1}{d_{p^\lambda}}\text{diag}(\mathbb{1}_{\sigma_\lambda})_{s,s}\mathbf{W}^\lambda_{s,t}
\end{equation}
\vspace{-0.3em}
for all $1 \leq s,t \leq n$. Here $\text{diag}(\mathbb{1}_{\sigma_\lambda})$ is a diagonal mask matrix representing the activation result, $\Psi$ is the set of all ($\Lambda$ + 1)-length random-walk paths on the
graph from node $v_h$ to $v_k$ , and $p^\lambda$ represents the $\lambda$-th node on a specific path p ($p^0$ and $p^\Lambda$ denote node i and k accordingly). 

Excluding one neighboring edge of node $v_h$ would bring two changes: (1) the degree of node $v_h$ will decrease to $d_h - 1$ as one of its neighbors is removed, and (2) There will be less random walk paths from node $v_h$ to $v_k$, $|\tilde{\Psi}|<|{\Psi}|$. 
Thus we have
\vspace{-0.2em}
\begin{equation}
    \resizebox{0.98\hsize}{!}{%
    $1- \mathbb{E}(\pdv{\tilde{x}_{h,s}^\Lambda}{x_{k,t}} / \pdv{x_{h,s}^\Lambda}{x_{k,t}}) = 1-\frac{\sqrt{(d_h-1)d_k} \sum_{p=1}^{\tilde{\Psi}}\mathbb{E}( \prod_{\lambda = 0}^{\Lambda}\frac{1}{d_{p^\lambda}}\text{diag}(\mathbb{1}_{\sigma_\lambda})_{s,s}\mathbf{W}^\lambda_{s,t})}{\sqrt{d_hd_k} \sum_{p=1}^{\Psi} \mathbb{E}(\prod_{\lambda=0}^{\Lambda}\frac{1}{d_{p^\lambda}}\text{diag}(\mathbb{1}_{\sigma_\lambda})_{s,s}\mathbf{W}^\lambda_{s,t})}$%
    }
    \label{eq: proof-gcn}
\end{equation}
\vspace{-0.2em}
From ~\cite{tang2020investigating}, we have  $\sum_{p=1}^{\Psi} \mathop{\mathbb{E}}(\prod_{\lambda = 0}^{\Lambda - 1} \frac{1}{d_p^\lambda} \text{diag}(\mathbb{1}_{\sigma_\lambda})_{s,s}\mathbf{W}^\lambda_{s,t}) = v$ is a constant.
Eq.~\ref{eq: proof-gcn} can rewritten as 
\begin{equation}
    \resizebox{0.98\hsize}{!}{%
    $1- \mathbb{E} (\pdv{\tilde{x}_{h,s}^\Lambda}{x_{k,t}} / \pdv{x_{h,s}^\Lambda}{x_{k,t}}) = 1-\sqrt{\frac{d_h-1}{d_h}}\frac{1/(d_h-1)\text{diag}(\mathbb{1}_{\sigma_\Lambda})_{s,s}\mathbf{W}^\Lambda_{s,t} \sum_{v_n \in \tilde{N}(h) } v }{1/(d_h)\text{diag}(\mathbb{1}_{\sigma_\Lambda})_{s,s}\mathbf{W}^\Lambda_{s,t} \sum_{v_n \in N(h) } v }$%
    }
\end{equation}
\vspace{-0.2em}
Then we have
\vspace{-0.7em}
\begin{equation}
\begin{split}
    1- \mathbb{E} (\pdv{\tilde{x}_{h,s}^\Lambda}{x_{k,t}} / \pdv{x_{h,s}^\Lambda}{x_{k,t}}) = 1- \sqrt{1 - \frac{1}{d_h}}
\end{split}
\end{equation}
Since if $d_h > d_l$, we can deduce $\sqrt{1-\frac{1}{d_h}} > \sqrt{1-\frac{1}{d_l}}$, thus $1- \mathbb{E}(\pdv{\tilde{x}_{h,s}^\Lambda}{x_{k,t}} / \pdv{x_{h,s}^\Lambda}{x_{k,t}}) < 1- \mathbb{E}(\pdv{\tilde{x}_{l,s}^\Lambda}{x_{k,t}} / \pdv{x_{j,s}^\Lambda}{x_{k,t}})$ and $D(k, h) < D(k, l)$ hold.

With the proof for GCN model, Theorem \ref{theorem} can be easily extended to general GNN models.
For general GNNs, the output node features of the $\Lambda$-th layer are generated as follows:
\vspace{-0.2em}
\begin{equation}
    x_h^{\Lambda+1} = \sigma(W^\Lambda\sum_{v_a \in N(h)} \alpha_{a,h}x_a^\Lambda)
\end{equation}
\vspace{-0.2em}
where $\alpha$ is a constant or parameters related with node attributes, such as node degrees or parameters will be learned, such as attention scores.
We calculate the effect of node $v_k$ on $v_h$ as follows:
\vspace{-0.5em}
\begin{equation}
    \mathbb{E} (\frac{\partial x_{h}^\Lambda}{\partial x_{k}}) = v d_k \sum_{v_n \in N(h)}\alpha_{n,h} \cdot \text{diag}(\mathbb{1}_{\sigma_\Lambda}) \cdot \mathbf{W}^\Lambda
\end{equation}
For the effect of node $v_k$ after excluding one target edge, the cardinality of the set of $v_h$ neighbor nodes decreases from $N$ to $N - 1$ and the value of $\alpha$ may also change to $\tilde{\alpha}$.
We have the effect ratio is
\vspace{-0.3em}
\begin{equation}
    \mathbb{E} (\frac{\partial \tilde{x_{h,s}}^\Lambda}{\partial x_{k,t}} / \frac{\partial x_{h,s}^\Lambda}{\partial x_{k,t}}) 
    = \frac{\sum_{v_n \in \tilde{N(h)}} \tilde{\alpha_{n,h}}}{\sum_{v_n \in N(h)}\alpha_{n,h}}
\end{equation}
If $\alpha$ is unrelated with the degree of $v_h$, then the theorem holds since $\alpha = \tilde{\alpha}$ the expectation of the ratio is $\frac{d_h - 1}{d_h}$ and $D(k, h) < D(k, l)$.
If the value of $\alpha \propto (d_h)^m$ then we have the raio is $\frac{(d_h - 1)(d_h - 1)^m}{d_h (d_h)^m}$. If $m >= -1$, we still have the theorem holds. To our best knowledge, we do not find the existing GNNs with $\alpha \propto (d_h)^m $ and $m < -1$, so our theorem holds for general GNNs.
\end{proof}

\end{document}